\documentclass{article}

\usepackage{arxiv}

\usepackage[utf8]{inputenc} 
\usepackage[T1]{fontenc}    
\usepackage{hyperref}       
\usepackage{url}            
\usepackage{booktabs}       
\usepackage{amsfonts}       
\usepackage{nicefrac}       
\usepackage{microtype}      
\usepackage{lipsum}
\usepackage{graphicx}
\usepackage{amsmath}
\usepackage{amssymb}
\usepackage{multirow}
\usepackage{float}
\usepackage{subfigure}

\newcommand{\tabincell}[2]{\begin{tabular}{@{}#1@{}}#2\end{tabular}}

\def\eg{\emph{e.g}.}
\def\etal{\emph{et al}.}

\title{PointShuffleNet: Learning Non-Euclidean Features with Homotopy Equivalence and Mutual Information}

\author{
 
  Linchao He \\
  Sichuan University\\
  \texttt{hlc@stu.scu.edu.cn} \\
   \And
  Mengting Luo \\
  Sichuan University\\
  \texttt{lmt@stu.scu.edu.cn} \\
  \And
  Dejun Zhang \\
  China University of Geosciences\\
  \texttt{zhangdejun@cug.edu.cn} \\
  \And
  Xiao Yang \\
  Sichuan University\\
  \texttt{xyang@scu.edu.cn} \\
  \And
  Hu Chen\footnote{Corresponding author} \\
  Sichuan University\\
  \texttt{huchen@scu.edu.cn} \\
  \And
  Yi Zhang\\
  Sichuan University\\
  \texttt{yzhang@scu.edu.cn} \\
}

\begin{document}
\maketitle
\begin{abstract}
Point cloud analysis is still a challenging task due to the disorder and sparsity of samplings of their geometric structures from 3D sensors. In this paper, we introduce the homotopy equivalence relation (HER) to make the neural networks learn the data distribution from a high-dimension manifold. A shuffle operation is adopted to construct HER for its randomness and zero-parameter. In addition, inspired by prior works, we propose a local mutual information regularizer (LMIR) to cut off the trivial path that leads to a classification error from HER. LMIR utilizes mutual information to measure the distance between the original feature and HER transformed feature and learns common features in a contrastive learning scheme. Thus, we combine HER and LMIR to give our model the ability to learn non-Euclidean features from a high-dimension manifold. This is named the non-Euclidean feature learner. Furthermore, we propose a new heuristics and efficiency point sampling algorithm named ClusterFPS to obtain approximate uniform sampling but at faster speed. ClusterFPS uses a cluster algorithm to divide a point cloud into several clusters and deploy the farthest point sampling algorithm on each cluster in parallel. By combining the above methods, we propose a novel point cloud analysis neural network called PointShuffleNet (PSN), which shows great promise in point cloud classification and segmentation. Extensive experiments show that our PSN achieves state-of-the-art results on ModelNet40, ShapeNet and S3DIS with high efficiency. Theoretically, we provide mathematical analysis toward understanding of what the data distribution HER has developed and why LMIR can drop the trivial path by maximizing mutual information implicitly.
\end{abstract}


\section{Introduction}
Point cloud analysis is challenging because it needs to process 3D data that is highly irregular and only contains Euclidean space information (\eg,~3D coordinates and normal vectors). The adjacent points may not be highly relevant and lack the connection information between points. Previous works used deep-learning-based methods to map the coordinates and other features (\eg,~normal vector and RGB information) to higher-dimension space to extract semantic information. As a result, point cloud analysis can be easily formulated as mining information from a Euclidean space problem. In this paper, we consider $\mathbb{R}^3$ space of point cloud as a submanifold, which is a local Euclidean space on a high-dimension non-Euclidean manifold. The data distribution of an object category should be a continuous closure subspace on the high-dimension Non-Euclidean manifold, and every point cloud that belongs to the above category should be contained in the subspace. Thus, we can formulate the point cloud analysis as a problem that learns the non-Euclidean features of the continuous closure subspaces from the corresponding discrete point clouds in $\mathbb{R}^3$.

To overcome the obstacle mentioned above, we propose a non-Euclidean feature learner (NEFL) enabling the transformation of point clouds into continuous closure subspaces efficiently. The non-Euclidean feature learner contains two parts: homotopy equivalence relation (HER) and the local mutual information regularizer (LMIR). The HER utilizes the homotopy equivalence transformation~\cite{3-manifolds2004Richard,Topology2000Anderson} to describe continuously deformed bijective relations between multiple objects in the same category (\textit{eg},~different chairs). Shuffle is the most efficient operation to perform homotopy transformation and can generate random and discrete submanifolds. There are multiple relations, called path-connected in topology, with the same endpoints on the manifold. Thus, we can transform a point cloud to other point clouds through different paths. Thus, the paths can be learned by the neural networks. This extends the generalization of the neural networks. We define the paths that connect several point clouds in the same category and are bundled in the continuous closure subspace of a same category as the non-trivial paths. Otherwise, we named the paths that are not bundled in a subspace as trivial paths. The trivial paths are hard to cut off explicitly on the high-dimension manifold, and this leads to poor generalization. Thus, we propose LMIR to cut off the trivial paths implicitly with mutual information estimation. LMIR estimates the mutual information between the original point clouds and HER-generated point clouds. Inspired by Deep INFOMAX~\cite{hjelm2019learning}, we use a contrastive loss that is based on contrastive loss to regularize our model to make the trivial paths have higher loss than the non-trivial paths. The LMIR can be deployed in the training phase to increase accuracy and can be removed in the inference phase to obtain higher speeds.

To further enhance the efficiency, we propose a parallel-friendly point sampling algorithm named ClusterFPS, which uses the divide-and-conquer method to divide multiple sampling regions and deploy the farthest points sampling (FPS) algorithm in each region. We implement a \textit{k}-means cluster sampling algorithm with a batch input as our dividing algorithm. The basic FPS algorithm is iterative and context sensitive. It is difficult to utilize modern GPU architectures to speed up the process. By using ClusterFPS as our sampling strategy, we can utilize the parallel computing of the GPUs to gain $5\times$ faster speed than FPS and comparable performance. 

Based on the non-Euclidean feature learner and ClusterFPS, we build a highly efficient neural network architecture named PointShuffleNet (PSN). Pointwise group convolution~\cite{zhang2018shufflenet,ma2018shufflenet} is introduced to replace MLP with better performance and fewer parameters. We modify the channel attention~\cite{hu2018senet} by concatenating a channel descriptor with the Euclidean coordinate of points. We achieve state-of-the-art performance on ModelNet~\cite{wu2015modelnet} and comparable results on ShapeNet~\cite{shapenet2015} and S3DIS~\cite{Armeni2017s3dis}. PSN achieves a $5\times$ speedup on various tasks. Remarkably, our model achieves the highest mean class accuracy ($91.6\%$) on ModelNet40 at 4.6 ms.

\section{Related Work}
\label{sec:headings}
Recently, deep-learning-based methods have been rapidly developed to process point clouds. They have shown great improvement in speed and accuracy.

\textbf{Pointwise MLP Methods.} PointNet~\cite{qi2016pointnet} was the first deep-learning method that used a sequence of multilayer perceptron (MLP) to directly processes point sets. PointNet showed great promises in accuracy but also weakness in model complexity and training speed. PointNet++~\cite{qi2017pointnetplusplus} follows the process in which convolutional neural networks (CNNs) extract information from local to global. It uses sampling and grouping layers to split point sets into small clusters and deploys PointNet to extract local features hierarchically. Compared with prior works, PointNet++ achieves better performance but is slower and more complicated. To further enhance the performance, Yang \etal~\cite{Yang2019PAT} proposed a pointwise-based method called Point Attention Transformers (PATs). This uses a parameter-efficient Group Shuffle Attention (GSA) to replace the complicated multihead attention in transformers. They also proposed a novel task-specific sampling method named Gumbel Subset Sampling (GSS). PointASNL~\cite{yan2020pointasnl} proposed a new adaptive sampling module to benefit the feature learning and avoid the biased effect of outliers. And they use local-nonlocal module to capture the neighbor and long-range dependencies of point clouds. PosPool~\cite{liu2020closerlook3d} proposed a local aggregation operator without weights named PosPool and combine it with a deep residual network achieving state-of-the-art results. Although pointwise MLP methods show great promise in efficiency, they still suffer from the inefficiency of FPS and lower accuracy than other approaches. Thus, we propose ClusterFPS and NEFL to further improve speed and accuracy.

\begin{figure}[!t]
	\begin{center}
		\includegraphics[width=0.45\linewidth]{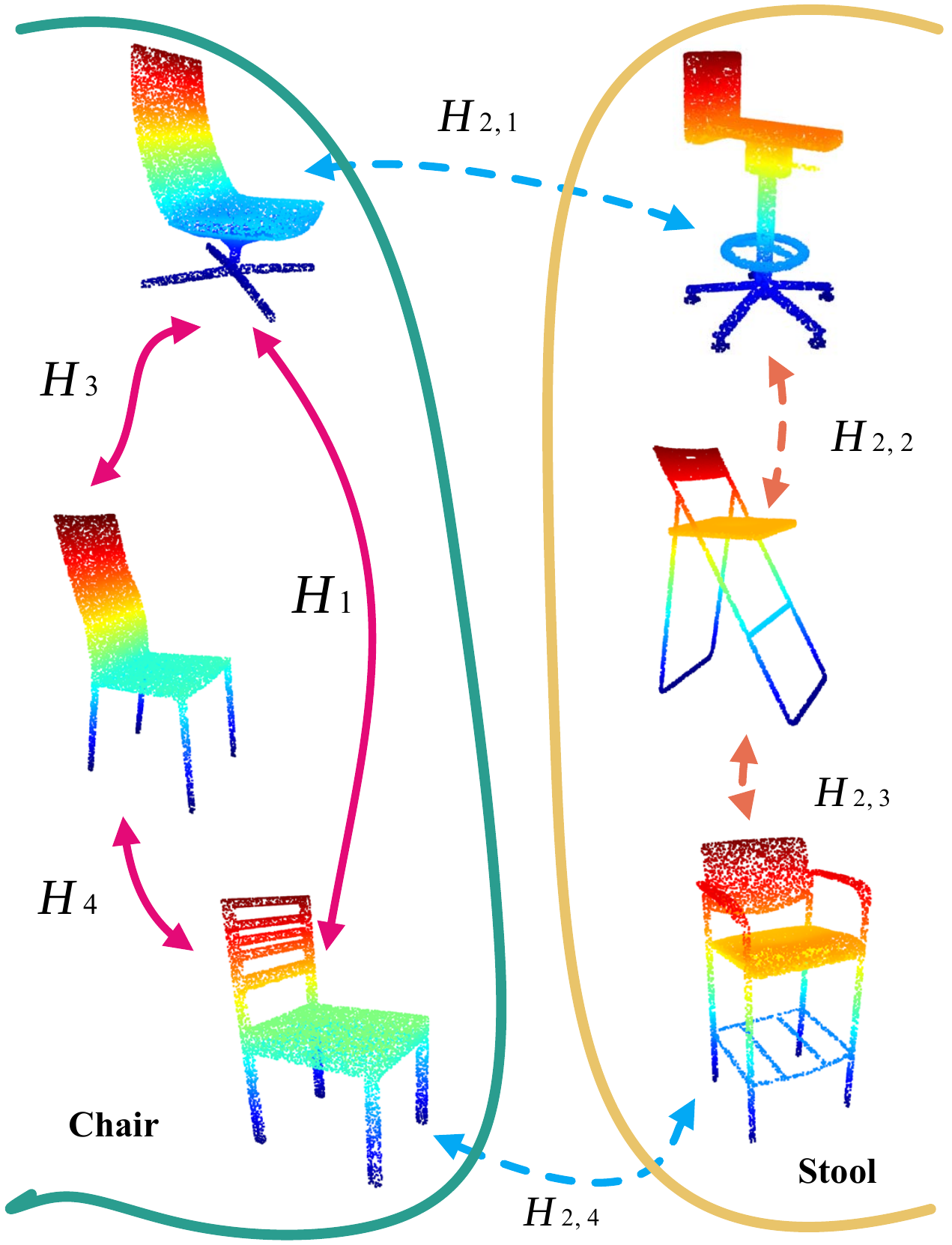}
	\end{center}
	\caption{Homotopy equivalence (\eg,~$\{H_1,H_3,H_4\}$ and $\{H_{2,1},H_{2,2},H_{2,3},H_{2,4}\}$) exists between objects. We can use homotopy equivalence relation (HER) to transform existing object to other objects, which is continuous mapping. Thus, neural network can capture above transformation and learn data distribution from it. However, there are several trivial paths (\eg,~$\{H_{2,1},H_{2,2},H_{2,3},H_{2,4}\}$) for chairs that may cause neural network to learn incorrect data distribution. It is difficult to cut off trivial path explicitly. Inspired by Deep INFOMAX~\cite{hjelm2019learning}, we propose local mutual information regularizer (LMIR) to perform mutual information estimation on $H$. Trivial paths (\eg,~$\{H_{2,1},H_{2,2}\}$) have less mutual information and higher loss than excepted path $\{H_1,H_3,H_4\}$ on chairs. However, $\{H_{2,1},H_{2,2}\}$ can be considered as standalone non-trivial paths for stools and lead neural network to have better generalization on stools.}
	\label{fig:nefl}
\end{figure}

\textbf{Convolution-based Methods.} CNNs have shown great success in image and video recognition, action analysis and natural language processing. Extending CNNs to process point cloud data has aroused wide interest from researchers. In PointConv~\cite{wu2018pointconv}, the convolutional operation is a Monte Carlo estimate of the continuous 3D convolution with an importance sampling. PointCNN~\cite{li2018pointcnn} achieves comparable accuracy by learning a local convolution order, but this has great weakness in convergence speed. RS-CNN~\cite{liu2019rscnn} uses 10-D hand crafted features as neighbor relationship and learn a dynamic convolution weights from the 10-D features. DensePoint~\cite{liu2019densepoint} utilizes dense connection mode to repeatedly aggregate different level and scale information in a deep hierarchy. KPConv~\cite{thomas2019KPConv} proposed a new point convolution method which processes radius neighborhoods with weights spatially located by a small set of kernel points. And they also develop a deformable version of KPConv that can fit the point cloud geometry. The follow-up work ShellNet~\cite{zhang-shellnet-iccv19} used statistics from concentric spherical shells to learn representative features, allowing the convolution to operate on feature spaces. However, the cost of downsampling points has become a speed bottleneck in processing large-scale point clouds.

\textbf{Graph-based Methods.} Simonovsky \etal~\cite{Simonovsky2017ecc} first considered each points as the vertex of a graph. They proposed Edge-Conditional Convolution (ECC), which uses a filter-generating network. Max pooling is utilized to aggregate the adjacency information. However, ECC has poor performance compared to pointwise-based and convolution-based networks. DGCNN~\cite{dgcnn} proposed a novel convolution layer named EdgeConv. DGCNN can construct a graph in the feature space and dynamically update the hierarchical structure. EdgeConv can capture local geometric features while ensuring the permutation invariant. While this achieves better results, DGCNN ignores the vector direction between adjacent points; this leads to some lost local geometric information. Lei \etal~\cite{lei2020spherical} proposed a novel graph convolution that uses a spherical kernel for 3D point clouds. Their spherical kernels quantize the local 3D space to geometric relationships. They built graph pyramids with range search and farthest point sampling, and named the whole network SPH3D-GCN. As a result, it is a challenge to design a neural network for point cloud analysis that needs to balance various factors such as model complexity, difficulty of implementation, accuracy and speed. However, graph-based methods need to construct graphs of points. This is ineffective and dynamic, and it is difficult to implement the networks and optimize the performance.

\section{Methods}
In this paper, we propose a new end-to-end pointwise neural network named PointShuffleNet (PSN) consisting of a non-Euclidean feature learner that can capture both Euclidean space and non-Euclidean space information, and ClusterFPS, which can downsample points uniformly and efficiently.

\subsection{Non-Euclidean Feature Learner}
\begin{figure}[!t]
	\centering
	\includegraphics[width=0.8\linewidth]{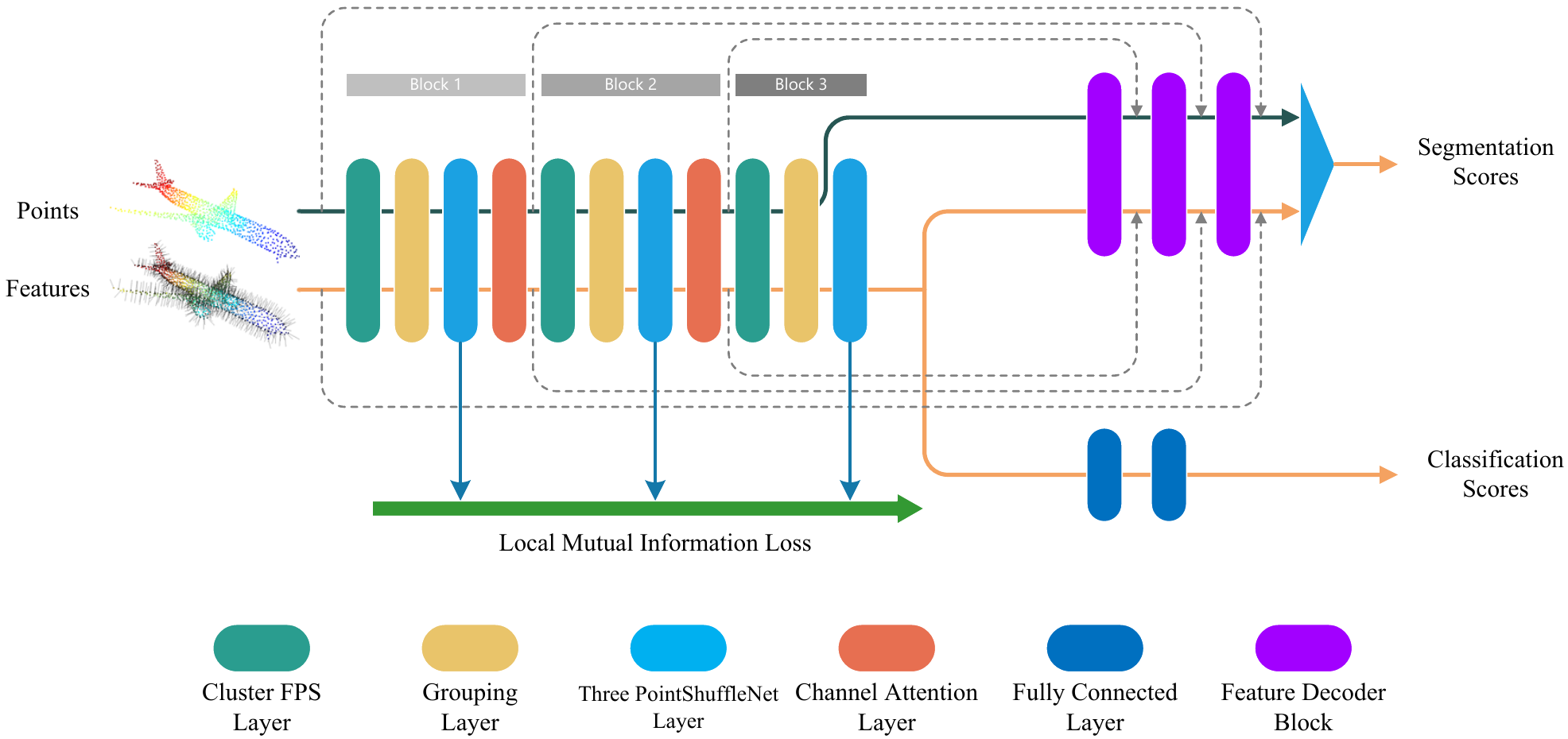}
	\caption{Illustration of PointShuffleNet for point cloud classification and segmentation. HER module is also used in feature decode layer.}
	\label{fig:psn}
\end{figure}

\label{sec:nefl}
As shown in Figure~\ref{fig:nefl}, the non-Euclidean feature learner aims to obtain better generalization for the neural network from the high-dimensional manifold and consists of two modules: 1) homotopy equivalence relation (HER) and 2) local mutual information regularizer (LMIR).
\subsubsection{Homotopy Equivalence Relation}
Given a set of point cloud $S = \{P_i^{\mathcal{C}} | i=1,\cdots,n\}$, we can consider $P_i^{\mathcal{C}}$ as a low-dimension embedding in $\mathbb{R}^3$ space of a higher-dimension manifold $M$ that belongs to a class $\mathcal{C}$. Thus, we can get a homotopy $H: P_i^{\mathcal{C}} \times [0,1] \to P_j^{\mathcal{C}}$ with a unit interval $[0,1]$, where homotopy $H$ contains infinite continuous bijective mappings. It is should be noticed that the homotopy $H$ between $P_i^{\mathcal{C}}$ and $P_j^{\mathcal{C}}$ are not unique. We can transform $P_i^{\mathcal{C}}$ to $ P_j^{\mathcal{C}}$ with infinite different paths. Therefore, we define a set $\mathcal{H} = \{H_i | i=1,2,\cdots,n\}$ that contains multiple homotopy paths of each point cloud pair (\eg,~$P_i^{\mathcal{C}}$ and $P_j^{\mathcal{C}}$).
\cite{basri2016efficient} showed that deep networks can efficiently extract the intrinsic, low-dimensional coordinates of a high-dimension. Thus, neural networks have the ability to learn the latent feature that represents the data distribution on the high-dimension manifold and project it into a low-dimension manifold. Thus, neural networks can learn the homotopy transformations $\mathcal{H}$, which are the data distribution on the high-dimension manifold, and classify $P_i^{\mathcal{C}}$ into class $\mathcal{C}$. The generalization of neural networks can be extended by constructing more effective homotopy equivalence relations on the input data. \cite{coetzee1995homotropy} constructed a homotopy equivalence function that deforms linear networks into nonlinear networks. Therefore, we can implement the homotopy equivalence function in single layer Perception (SLP) and multilayer perception (MLP). However, SLP and MLP both suffer from additional computational costs and have small coverage over the data distribution.

To overcome the above obstacle, a shuffle operation is introduced in this paper as a zero-overhead homotopy equivalence transformation. \cite{3-manifolds2004Richard,Topology2000Anderson} proved that the modulo shuffle is a homotopy equivalence in which the manifold is glued together along primitive solid torus components of its characteristic submanifold. Compared to SLP and MLP, modulo shuffle is parameter-free and can be implemented by reading data with a fixed stride from memory. A module shuffle can generate pseudo submanifolds from the known data distribution and improve the generalization ability of the model. 

In this paper, we propose two shuffle-based functions: sample shuffle and channel shuffle.

\textit{Sample Shuffle. } Since large-scale point clouds are going to be processed hierarchically, it is critical to sample and group information from neighbor points. The sample shuffle can construct a local homotopy equivalence transformation. For the sake of clarity, we call $p$ and $f$ the points from point cloud $P \in \mathbb{R}^{N \times 3}$ and their corresponding features from $F \in \mathbb{R}^{N \times D}$ respectively. We can concatenate each $f$ with corresponding $p$ to low-dimension manifolds $m \in \mathbb{R}^{N \times (3+D)}$, which is defined as
\begin{equation}
	m = p \oplus f
\end{equation}
where $\oplus$ is the concatenation operation. The sample shuffle function $\mathcal{S}_\mathit{sample}(\cdot)$ can be efficiently and elegantly implemented by reshaping, transposing and flattening~\cite{zhang2018shufflenet}. Thus, we can integrate sample shuffle into concatenate and define this as
\begin{equation}
	\hat{m} = p \oplus \mathcal{S}_\mathit{sample}(f)
	\label{eq:sample_shuffle}
\end{equation}
Since each point $p$ does not correspond to the original feature $f$, it is obvious that $\hat{m}$ and $m$ satisfy the homotopy equivalence on the high-dimension manifold.

\textit{Channel Shuffle. } \cite{zhang2018shufflenet} proposed the channel shuffle operation, which showed great promise for mobile devices in image classification. However, Zhang~\etal did not explain this mathematically in their paper. The channel shuffle operation can be considered an example of HER that operates on the channel dimension. ShuffleNet uses channel shuffle on the output of group convolution to approximate the output distribution of a non-group convolution layer. Thus, we combine sample shuffle and channel sample as
\begin{equation}
	\mathcal{S}_\psi(f, p) = \mathcal{S}_\mathit{channel}(\sigma_\psi (p \oplus \mathcal{S}_\mathit{sample}(f)))
\end{equation}
where $\sigma_\psi$ is a learnable mapping function (\eg, MLP) with parameter $\psi$ that maps $\hat{m}$ into high-dimension space. Eventually, we can ensure that the homotopy equivalence relation $\mathcal{S}_\psi(f, p)$ can help the model to be more general and have zero cost on computation. We will conduct an ablation study to show the performance of both shuffle operations in Section~\ref{sec:ablation}.

\subsubsection{Local Mutual Information Regularizer}
Although the homotopy equivalence relation $\mathcal{S}_\psi(f, p)$ can improve the generalization performance of the neural network models, it may suffer from bad generalization caused by trivial homotopy equivalence path $\hat{H}$. Due to the high complexity of high-dimensional space, it is difficult to cut off $\hat{H}$ explicitly. Therefore, we propose a novel approach named the local mutual information regularizer (LMIR) to cut off trivial path $\hat{H}$ implicitly. Let us now focus on the original feature $X$ and mapped feature $Z = \sigma_\psi(X)$. We define a mutual information function between $X$ and $Z$ as
\begin{equation}
	I(X,Z)=\iint p(z|x)\Tilde{p}(x)\log\frac{p(z|x)}{p(z)}dxdz
\end{equation}
where $x=m$, $\Tilde{p}(x)$ is the original distribution of data and $p(z)=\int p(z|x)\Tilde{p}(x)dx$. In~\cite{hjelm2019learning}, the researchers proposed a novel approach named Deep INFOMAX (DIM) to maximize mutual information between the input and output. Deep INFOMAX shows great promise in unsupervised learning. Deep INFOMAX estimates and maximizes the mutual information in one architecture and finds the most discriminative feature of input. DIM follows Mutual Information Neural Estimation (MINE)~\cite{belghazi2018mine} to estimate and replace the KL divergence estimator with a non-KL divergence estimator (\eg, Jensen-Shannon divergence estimator~\cite{nowozin2016fgan}) as
\begin{equation}
	\begin{split}
		\hat{I}^{(JS)}_{w,\phi}(X,\sigma_\psi(X),\mathcal{S}_\psi(\hat{X})) := \mathbb{E}&_\mathbb{P}[-\Phi(-T_w(x,\sigma_\psi(x)))] \\
		- \mathbb{E}&_{\tilde{\mathbb{P}}}[\Phi~(T_w(x,\mathcal{S}_\psi~(\hat{x}))]
	\end{split}
	\label{Eq:js}
\end{equation}
where $T_\omega$ is a weight-sharing discriminator with parameters $w$, $\mathbb{P}$ denotes the distribution of $X$, $\tilde{\mathbb{P}}=\mathbb{P}$, $X = \hat{X}$ and $\Phi(x)=\log(1+e^x)$ is the softplus function. Deep INFOMAX can maximize the above equation in a contrastive learning scheme. Deep INFOMAX finds the discriminative information between samples and is not concerned with the classes of samples. It shows great promise in unsupervised learning. However, Deep INFOMAX cannot increase the performance in our supervised task (\eg,~classification and segmentation) because it is only concerned with the discriminative information between samples and ignores the common features of samples that are in the same category. And Deep INFOMAX only considers the shuffled feature as a negative sample. In our theory, we can treat the shuffled feature as a positive sample if the path is non-trivial. Thus, we modify Equation~\ref{Eq:js} as
\begin{equation}
	\label{Eq:lmir}
	\begin{split}
		\Tilde{I}^{(JS)}_{w,\phi}(X,\sigma_\psi(X),\mathcal{S}_\psi(\hat{X})) := \mathbb{E}&_{\tilde{\mathbb{P}}}[-\Phi(-T_w(x,\mathcal{S}_\psi(\hat{x})))] \\ - \mathbb{E}&_\mathbb{P}[\Phi(T_w(x,\sigma_\psi(x))]
	\end{split}
\end{equation}

\begin{figure}[!t]
	\begin{center}
		\includegraphics[width=0.45\linewidth]{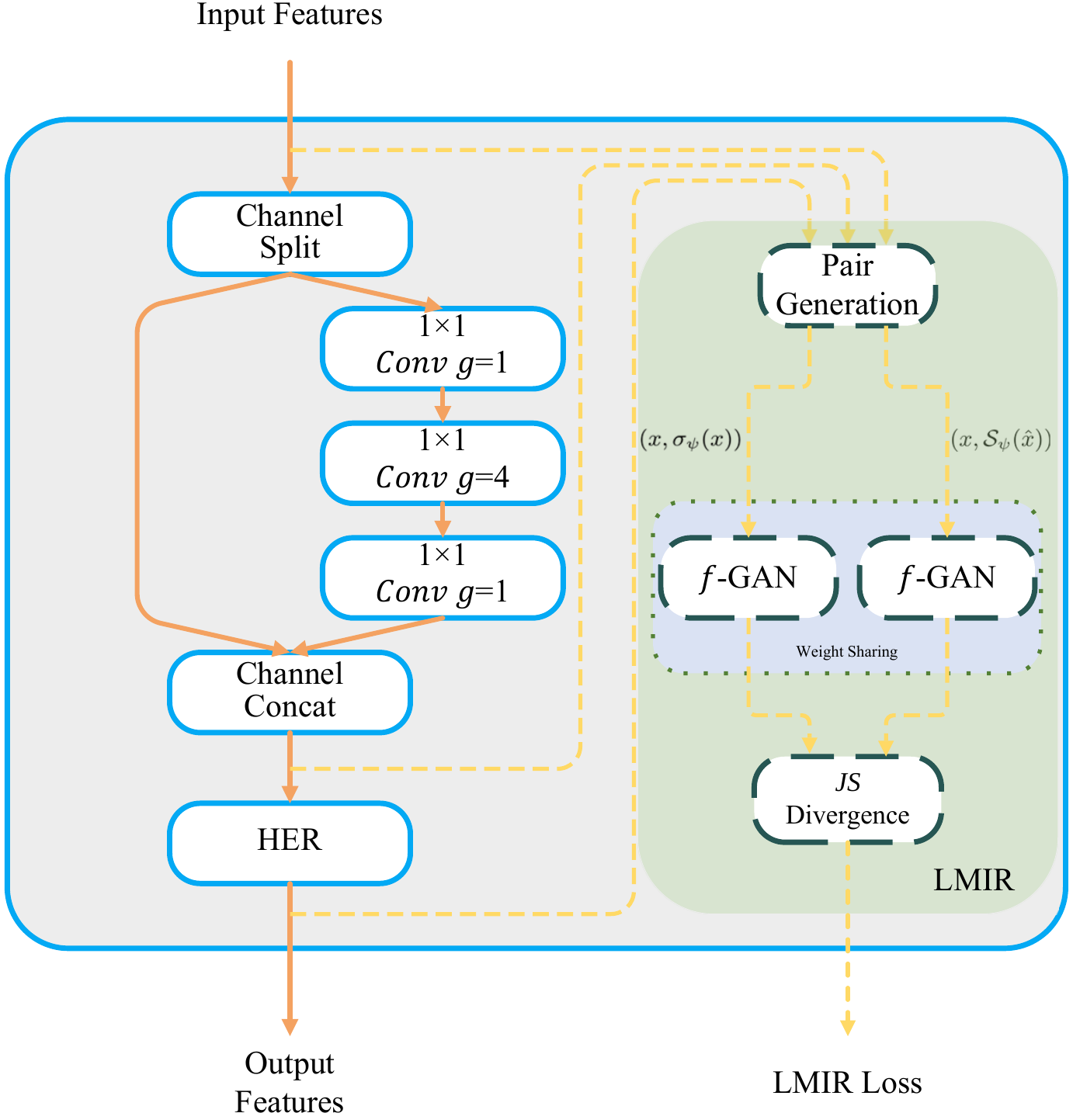}
	\end{center}
	\caption{Structure of PointShuffleNet layer. Orange lines and blue boxes denote forward propagation of input feature. We can sample $\sigma_\psi(x)$ and $\mathcal{S}_\psi(\hat{x})$ after the channel concatenation operation and HER module. Thus, we can get two feature pairs $(x,\sigma_\psi(x))$ and $(x,\mathcal{S}_\psi(\hat{x}))$ by pair generation operation. Mutual information estimated pairs are processed by LMIR to generate LMIR loss. Dotted line and dotted frame indicate these be deleted in inference phase.}
	\label{fig:psnlayer}
\end{figure}

We exchange $(x,\mathcal{S}_\psi(\hat{x}))$ and $(x,\sigma_\psi(x))$ to get a new mutual information estimator named the local mutual information regularizer (LMIR). LMIR can distinguish whether the feature is from a non-trivial path through mutual information and guides the model to learn the more similar shuffled features and punishes the less similar shuffled features which are likely to share less mutual information with the original features.It constrains the neural network to learn non-trivial homotopy equivalence path $H$ inside the same category instead of trivial homotopy equivalence path $\hat{H}$ between categories. We then define the regularization loss of our LMIR on the points, maximizing the average estimated MI:
\begin{equation}
	\begin{split}
		L_{LMIR}=\frac{1}{C}\sum^{C}_{i=1}\Tilde{I}^{(JS)}_{w^{(i)},\phi^{(i)}}(X,\sigma_{\psi^{(i)}}(X),\mathcal{S}_{\psi^{(i)}}(\hat{X}))
	\end{split}
\end{equation}
where $C$ is the number of PointShuffleNet layers (its implementation details are provided in Section~\ref{sec:psn}). LMIR can be implemented by simply attaching a two-layer discriminator $T_{w^{(i)}}$ after every HER module with some parameters increasing. However, LMIR can be excluded from the inference phase without losing accuracy because LMIR is not a part of feature generation.

\subsection{ClusterFPS}
\label{sec:clusterfps}
In the original farthest point sampling implementation, the algorithm sample $K$ points from a point cloud $P$ with $N$ points and returns a downsampling of the metric space $\hat{P} = \{\hat{p}_1,\hat{p}_2,\cdots,\hat{p}_K\}$ in which each $\hat{p}_k$ is the farthest point from the first $k-1$ points. Although it has a good coverage of the point cloud, it is obvious that the algorithm needs to know the position information of the previous points before sampling a new point. This is difficult to speed up through parallel computing. According to~\cite{hu2019randla}, FPS takes up to 200 seconds to sample $10\%$ of $10^6$ points.

To address the above issue, we propose a parallel-friendly sampling algorithm named ClusterFPS that aims to ensure good coverage of the point cloud as quickly as possible. 

\textit{Computing Cluster Centers.}
For a given point cloud $P$, its $n$ cluster $\{c_1,c_2,\cdots,c_n\}$ can be computed by the efficient $k$-means algorithm. A standard $k$-means algorithm is unable to take batch data as input. Thus, we implement a parallel version of the $k$-means algorithm to utilize the GPUs.

\textit{Finding Neighboring Points. } 
For each $c_i$ cluster, we use $K$-Nearest Neighbors ($k$-NN) to query its neighboring points. We sample $m$ nearest points from $P$ to get a set of points $P_{c_i} = \{p^{c_i}_1,p^{c_i}_2,\cdots,p^{c_i}_m\}$.

\textit{Parallel Farthest Point Sampling. }
For each $c_i$ cluster, a subset $\hat{P}_{c_i}=\{\hat{p}^{c_i}_1,\hat{p}^{c_1}_2,\cdots,\hat{p}^{c_i}_r\}$ of points $P_{c_i}$ can be sampled by FPS in parallel.

\textit{Grouping Cluster Points}
Eventually, the output of parallel farthest point sampling is sets of downsampled points $\hat{P}_{c_i}$ in which each set contains $r$ sampled points. Thus, we can obtain the final downsampled point set $\hat{P} = \{\hat{p}^{c_1}_1,\hat{p}^{c_1}_2,\cdots,\hat{p}^{c_1}_r,\cdots,\hat{p}^{c_n}_r\}$ by grouping $\hat{P}_{c_i}$ into one set.

Overall, our ClusterFPS algorithm is designed to reduce context dependencies and run in parallel. Thus, this algorithm can utilize GPUs to compute more quickly. This is discussed in Section~\ref{sec:speed}.

\section{PointShuffleNet}
\label{sec:psn}

By combining the above three components proposed in Section~\ref{sec:clusterfps} and Section~\ref{sec:nefl}, we implement a hierarchical neural network for both classification and segmentation tasks as shown in Figure~\ref{fig:psn}. We combine the cross-entropy loss with LMIR regularization loss as our loss function.

As shown in Figure~\ref{fig:psnlayer}, we build a basic PointShuffleNet layer for both classification and segmentation tasks. It uses features as input, and its channel is split into two groups to reduce the computation complexity~\cite{ma2018shufflenet}. Then, a 3-layer $1\times1$ convolution is utilized to extract features and to be concatenated with residual features. The output features $\mathcal{S}_\psi(\hat{x})$ can be obtained by HER from $\sigma_\psi(x)$. We sample $x$, $\sigma_\psi(x)$ and $\mathcal{S}_\psi(\hat{x})$ from the mainstream of layer (orange line) to construct pairs: $(x,\mathcal{S}_\psi(\hat{x}))$ and $(x,\sigma_\psi(x))$. The LMIR module computes the regularization loss from the outputs of the $f$-GAN discriminator. The dotted yellow line and dotted black frame indicate these can be deleted in the inference phase.

For the classification task, we designed a three-block feature extractor followed by a classifier. Each block contains a sampling layer (ClusterFPS), grouping layer (same as~\cite{qi2017pointnetplusplus}) and three PointShuffleNet layers. The first two blocks sample 512 and 256 points, and a max-pooling layer is adopted to aggregate local features. The last block aggregates the final features from all remaining points. The final classification scores can be computed by a classifier with fully connected layers, dropout and softmax activation. Each $1\times1$ convolutional layer is followed by a batch normalization layer and the ReLU activation function. Channel attention is also deployed to enhance the representation between blocks.

\begin{table}[!t]
	\centering
	\tabcolsep=0.07cm
	\begin{tabular}{c|cccccc}
		\hline
		Method & Input & Points & Params & Class & OA & Infer(ms) \\
		\hline
		PointNet~\cite{qi2016pointnet} & P & 1k & 3.5M & 86.2 & 89.2 & 2.5 \\
		SO-Net~\cite{li2018sonet} & P & 2k & 2.4M & 87.3 & 90.9 &  - \\		
		SPH3D~\cite{lei2020spherical} & P & 10k & 0.8M & 89.3 & 92.1 & 8.4 \\
		DGCNN~\cite{dgcnn} & P & 1k & 1.8M & 90.2 & 92.2 & 5.6 \\
		PointCNN~\cite{li2018pointcnn} & P & 1k & \textbf{0.6M} & - & 92.2 & 7.5 \\
		KPConv~\cite{thomas2019KPConv} & P & 6.8k & 14.3M & - & 92.9 & 21.5 \\	
		PointASNL~\cite{yan2020pointasnl} & P & 1k & - & - & 92.9 & - \\
		Grid-GCN~\cite{Xu2020Grid-GCN} & P & 1k & - & 91.3 & 93.1 & 2.6 \\
		PosPool~\cite{liu2020closerlook3d} & P & 10k & 19.4M & - & 93.2 & - \\
		RS-CNN~\cite{liu2019rscnn} & P & 1k & 1.4M & - & 93.6 & 4.3 \\
		\hline	
		PointNet++~\cite{qi2017pointnetplusplus} & P,N & 5k & 1.5M & 90.7 & 91.9 & \textbf{1.3} \\
		PAT~\cite{Yang2019PAT} & P,N & 1k & 0.6M & - & 91.7 &  11 \\
		SO-Net~\cite{li2018sonet} & P,N & 5k & 2.4M & 89.3 & 92.3 &  - \\
		SpiderCNN~\cite{xu2018spidercnn} & P,N & 5k & - &- & 92.4 & - \\
		A-CNN~\cite{komarichev2019acnn} & P,N & 1k & -  & 90.3 & 92.6 & - \\
		PointASNL~\cite{yan2020pointasnl} & P,N & 1k & - & - & 93.2 & - \\
		\hline
		PSN & P & 1k & 1.4M & 90.5 & 92.7 & $4.6^*$ \\
		PSN & P,N & 1k & 1.4M & \textbf{91.6} & \textbf{93.2} & $4.6^*$ \\		
		\hline
	\end{tabular}
	\caption{Mean class accuracy and overall accuracy on ModelNet40. ``Params'' stands for number of parameters. ``$*$'' stands for inference without LMIR. ``Train'' denotes forward and backward propagation time per sample, and ``Infer'' denotes forward propagation per sample. ``P'' stands for coordinates for point and ``N'' stands for normal vector.}
	\label{tab:classifiy}
\end{table}

\begin{table}[!t]
	\centering
	\tabcolsep=0.06cm
	\begin{tabular}{c|ccccc}
		\hline
		Method & Size(m) & Points & mIoU & OA & Infer(ms) \\ \hline
		PointCNN~\cite{li2018pointcnn} & $1.5\times1.5 $ &4096 & 57.26 & 85.91 & - \\
		Grid-GCN~\cite{Xu2020Grid-GCN} & $1.5\times1.5 $ &4096 & 57.75 & 86.94 & 25.9 \\
		PAT~\cite{Yang2019PAT} & $1.5\times 1.5 $ &2048 & 64.3 & - & - \\
		PointASNL~\cite{yan2020pointasnl} & $1.5 \times 1.5 $ &8192 & \textbf{68.7} & \textbf{87.7} & - \\
		\hline 
		
		PointNet~\cite{qi2016pointnet} & $1\times1$ & 4096 & 41.09 & - & 20.9 \\
		DGCNN~\cite{dgcnn} & $1\times1$ & 4096 & 47.94 & 83.64 & 178.1 \\
		PointNet++~\cite{qi2017pointnetplusplus} & $1\times 1$ &4096 & 53.2 & - & - \\
		\hline
		PSN & $1 \times 1 $ & 4096 & \textbf{55.2} & \textbf{86.34} & 28.3 \\ \hline
	\end{tabular}
	\caption{Mean class IoU and overall accuracy on indoor S3DIS dataset. ``Infer'' denotes forward propagation time per sample.}
	\label{tab:s3dis}
\end{table}

\begin{table*}[!t]
	\centering
	\tabcolsep=0.03cm
	\begin{tabular}{c|cc|cccccccccccccccc}
		\hline
		Method &
		\tabincell{c}{instance\\mIoU} &
		\tabincell{c}{class\\mIoU} &
		\tabincell{c}{air\\plane} &
		bag &
		cap &
		car &
		chair &
		\tabincell{c}{ear\\phone} &
		guitar &
		knife &
		lamp &
		laptop &
		\tabincell{c}{motor\\bike} &
		mug &
		pistol &
		rocket &
		\tabincell{c}{Skate\\board} &
		table \\
		\hline
		\multicolumn{3}{c|}{Number} & 2690 & 76 & 55 & 898 & 3758 & 69 & 787 & 392 & 1547 & 451 & 202 & 184 & 286 & 66 & 152 & 5271 \\
		\hline
		PointNet~\cite{qi2016pointnet} & 83.7 & 80.4 & 83.4 & 78.7 & 82.5 & 74.9 & 89.6 & 73.0 & 91.5 & 85.9 & 80.8 & 95.3 & 65.2 & 93.0 & 81.2 & 57.9 & 72.8 & 80.6 \\
		SO-Net~\cite{li2018sonet} & 84.9 & 81.0 & 82.8 & 77.8 & 88.0 & 77.3 & 90.6 & 73.5 & 90.7 & 83.9 & 82.8 & 94.8 & 69.1 & 94.2 & 80.9 & 53.1 & 72.9 & 83.0 \\
		PointNet++~\cite{li2018sonet} & 85.1 & 81.9 & 82.4 & 79.0 & 87.7 & 77.3 & 90.8 & 71.8 & 91.0 & 85.9 & 83.7 & 95.3 & 71.6 & 94.1 & 81.3 & 58.7 & 76.4 & 82.6 \\
		DGCNN~\cite{dgcnn} & 85.1 & 82.3 & 84.2 & 83.7 & 84.4 & 77.1 & 90.9 & 78.5 & 91.5 & 87.3 & 82.9 & 96.0 & 67.8 & 93.3 & 82.6 & 59.7 & 75.5 & 82.0 \\
		P2Sequence~\cite{liu2019point2sequence} & 85.2 & 82.2 & 82.6 & 81.8 & 87.5 & 77.3 & 90.8 & 77.1 & 91.1 & 86.9 & 83.9 & 95.7 & 70.8 & 94.6 & 79.3 & 58.1 & 75.2 & 82.8 \\
		PointCNN~\cite{li2018pointcnn} & 86.1 & 84.6 & 84.1 & 86.5 & 86.0 & 80.8 & 90.6 & 79.7 & 92.3 & 88.4 & 85.3 & 96.1 & 77.2 & 95.2 & 84.2 & 64.2 & 80.0 & 83.0 \\
		RS-CNN~\cite{liu2019rscnn} & 86.2 & 84.0 & 83.5 & 84.8 & 88.8 & 79.6 & 91.2 & 81.1 & 91.6 & 88.4 & 86.0 & 96.0 & 73.7 & 94.1 & 83.4 & 60.5 & 77.7 & 83.6 \\
		PointASNL~\cite{yan2020pointasnl} & 86.1 & 83.4 & 84.1 & 84.7 & 87.9 & 79.7 & 92.2 & 73.7 & 91.0 & 87.2 & 84.2 & 95.8 & 74.4 & 95.2 & 81.0 & 63.0 & 76.3 & 83.2 \\
		\hline
		PSN & 85.8 & 82.5 & 83.5 & 81.4 & 87.9 & 78.8 & 91.1 & 74.5 & 90.6 & 87.1 & 84.9 & 95.8 & 71.6 & 95.2 & 81.0 & 57.8 & 75.7 & 83.8 \\
		\hline
	\end{tabular}
	\caption{Mean instance IoU and mean class IoU on part segmentation ShapeNet dataset. Per-class IoU is also illustrated.}
	\label{tab:shapenet_detail}
\end{table*}

For the segmentation task, the configuration of the extractor is similar to the configuration in the classification task. Following~\cite{qi2017pointnetplusplus}, we construct three blocks as decoder. Each block consists of distance-based interpolation, across-level skip links and MLP with HER. 

\section{Experiments}

\begin{table*}[t]
	\centering
	\tabcolsep=0.06cm
	\begin{tabular}{c|ccc|ccccccccccccc}
		\hline
		Method & mACC & mIoU & OA & ceiling & floor & wall & beam & column & window & door & table & chair & sofa & bookcase & board & clutter \\
		\hline
		PointNet~\cite{qi2016pointnet} & 49.0 & 41.1 & - & 88.8 & 97.3 & 69.8 & 0.1 & 3.9 & 46.3 & 10.8 & 52.6 & 58.9 & 40.3 & 5.9 & 26.4 & 33.2 \\
		PointCNN~\cite{li2018pointcnn} & 63.9 & 57.3 & 85.9 & 92.3 & 98.2 & 79.4 & 0.0 & 17.6 & 22.8 & 62.1 & 74.4 & 80.6 & 31.7 & 66.7 & 62.1 & 56.7 \\
		PointWeb~\cite{zhao2019pointweb} & 66.6 & 60.3 & 87.0 & 92.0 & 98.5 & 79.4 & 0.0 & 21.1 & 59.7 & 34.8 & 76.3 & 88.3 & 46.9 & 69.3 & 64.9 & 52.5 \\
		HPEIN~\cite{jiang2019hpein} & 68.3 & 61.9 & 87.2 & 91.5 & 98.2 & 81.4 & 0.0 & 23.3 & 65.3 & 40.0 & 75.5 & 87.7 & 58.5 & 67.8 & 65.6 & 49.7 \\
		PointASNL~\cite{yan2020pointasnl} & 68.5 & 62.6 & 87.7 & 94.3 & 98.4 & 79.1 & 0.0 & 26.7 & 55.2 & 66.2 & 83.3 & 86.8 & 47.6 & 68.3 & 56.4 & 52.1 \\
		\hline
		PSN & 63.85 & 55.2 & 86.3 & 91.5 & 98.2 & 74.2 & 0.0 & 6.0 & 54.7 & 22.4 & 73.6 & 79.3 & 48.2 & 60.3 & 57.5 & 51.4 \\
		\hline
	\end{tabular}
	\caption{Mean instance IoU, mean class IoU and overall accuracy on indoor semantic segmentation S3DIS dataset. Per-class IoU is also shown.}
	\label{tab:s3dis_detail}
\end{table*}

We evaluate PointShuffleNet on several widely used datasets including ModelNet40~\cite{wu2015modelnet}, ShapeNet~\cite{shapenet2015} and S3DIS~\cite{Armeni2017s3dis}. To demonstrate the efficiency of our method, we also report speed and performance in a fair comparison. All experiments are conducted in PyTorch~\cite{pytorch} on a single NVIDIA GTX 1080ti GPU with a 11 GB VRAM.
We optimize the networks by using Adam~\cite{kingma2014adam} with an initial learning rate of 0.001 and batch size of 24. We deploy a cosine-annealing decay schedule~\cite{loshchilovH2017sgdr} on the learning rate with a period $T_{max}=32$.

\subsection{Point Cloud Classification}
The Modelnet40 dataset~\cite{wu2015modelnet} includes 12,311 CAD models from 40 classes. Following the official settings, we use 9,843 models to train our network and 2,468 models to evaluate. Following the configuration of~\cite{qi2016pointnet,qi2017pointnetplusplus}, we sample 1024 points uniformly from raw points and compute the normal vectors from corresponding mesh models. We also use data augmentation in the same way  as~\cite{qi2016pointnet,qi2017pointnetplusplus} by randomly rotating the data along the $z$-axis and jittering each point with a Gaussian that has zero mean and 0.02 standard deviation.

Classification results on the test set are summarized in Table~\ref{tab:classifiy}. Our PointShuffleNet achieves the state-of-the-art accuracy on ModelNet40 except RS-CNN~\cite{liu2019rscnn}. RS-CNN improves their results from $92.2\%$ to $93.6\%$ with multiple abstraction scales and tricky voting test strategy. Their training and evaluate protocols are different from other normal protocols. Compared to other methods implemented in TensorFlow with better GPUs, our method still obtains a great balance between speed and accuracy. An analysis of the running time will be demonstrated in Section~\ref{sec:speed}. It is remarkable that LMIR makes PSN have the highest mean class accuracy due to its great generalization on each class.

\subsection{Point Cloud Segmentation}

\label{sec:speed}

We evaluate our PSN on two larger-scale segmentation tasks: ShapeNet~\cite{shapenet2015} for part segmentation and Stanford 3D Large-Scale Indoor Spaces (S3DIS)~\cite{Armeni2017s3dis} for indoor scene segmentation. The ShapeNet dataset contains 16,881 shapes (14,006 models for training and 2,874 for evaluating) from classes and has 50 parts in total. We follow the configuration of PointNet++~\cite{qi2017pointnetplusplus} using 2048 points as our input. There are 6 large-scale indoor areas with 271 rooms in the S3DIS dataset, and each point is annotated with one of 13 categories. Because area 5 is the only area that has no overlap with other areas, we consider areas 1-4 and 6 as our training split and area 5 as our evaluation split. Following~\cite{yan2020pointasnl}, we generate training data by randomly sampling rooms into $1m\times1m$ with 4096 points during the training process. At test time, we split the rooms into $1m\times1m$ blocks with 4096 points and stride $0.5m$. 

In Table~\ref{tab:shapenet_detail}, we show the quantitative results of PSN on ShapeNet with other state-of-the-art methods~\cite{qi2016pointnet,li2018sonet,qi2017pointnetplusplus,dgcnn,liu2019point2sequence,li2018pointcnn,liu2019rscnn,yan2020pointasnl} under the same training and evaluating protocols. PSN shows comparable results with other methods. Compared with PointNet~\cite{qi2016pointnet} and PointNet++~\cite{qi2017pointnetplusplus}, we have notably increased both mIoU and class mIoU due to the effectiveness of NEFL.

The evaluation performance of S3DIS dataset on Area 5 is shown in Table~\ref{tab:s3dis} and Table~\ref{tab:s3dis_detail}. Our method achieves comparable results with a block size of $1m\times1m$ and point input of $4096$.  Our method achieves better results with a block size of $1m\times1m$ and point input of $4096$ compared with previous methods which use same experiment setting. And our method also achieves comparable performance compared with other methods which use bigger areas or more points as input.


\subsection{Ablation Study}
\label{sec:ablation}
\begin{table*}[!htbp]
	\centering
	\begin{tabular}{c|c|c|cccc|cc}
		\hline &
		\multicolumn{1}{c|}{\multirow{2}{*}{Model}} &
		\multicolumn{1}{c|}{\multirow{2}{*}{Points}} &
		\multicolumn{1}{c}{\multirow{2}{*}{Normal}} &
		\multicolumn{2}{|c|}{NEFL} &
		\multicolumn{1}{c|}{\multirow{2}{*}{ClusterFPS}} &
		\multicolumn{2}{c}{Accuracy} \\ \cline{5-6} \cline{8-9} &
		\multicolumn{1}{c|}{} &
		\multicolumn{1}{c|}{} &
		\multicolumn{1}{c}{} &
		\multicolumn{1}{|c|}{HER} &
		\multicolumn{1}{|c|}{LMIR} &
		\multicolumn{1}{c|}{} &
		\multicolumn{1}{c}{class} &
		instance \\ \hline
		A & PointNet++ & 1024 & & & & & 88.0 & 90.7 \\
		B & PointNet++ & 5000 & \checkmark & & & & - & 91.9 \\
		C & PointNet++ & 1024 & \checkmark & \checkmark & & & 91.0 & 92.5 \\ \hline
		D & PSN & 1024 & \checkmark & & & \checkmark & 90.9 & 92.4 \\
		E & PSN & 1024 & \checkmark & \checkmark & & \checkmark & 91.1 & 92.8 \\
		F & PSN & 1024 & \checkmark & \checkmark & $*$ & \checkmark & 90.9 & 92.9 \\
		G & PSN & 1024 & \checkmark & \checkmark & \checkmark &  & 91.5 & 93.0 \\
		H & PSN & 1024 & \checkmark & \checkmark & \checkmark & \checkmark & 91.6 & 93.2 \\ \hline
	\end{tabular}
	\caption{Ablation stud on ModelNet40. $*$ denotes that we use original local mutual information loss from \cite{hjelm2019learning} to replace our LMIR loss.}
	\label{tab:ablation}
\end{table*}
\begin{figure}
	\centering
	\begin{minipage}[c]{.5\textwidth}
		\centering
		\includegraphics[width=0.7\linewidth]{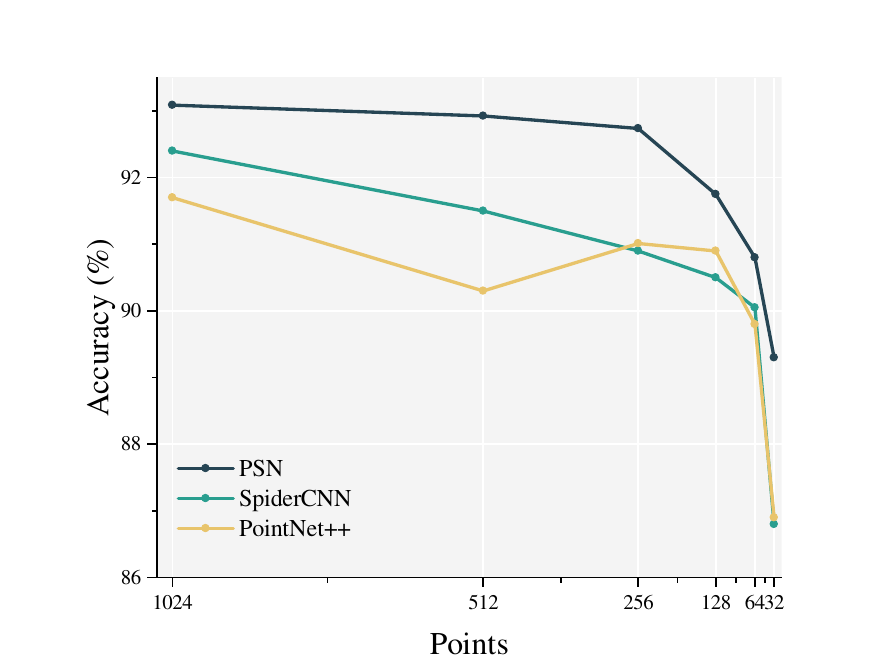}
		\caption{Train and evaluate on sparse input.} 
		\label{fig:train_sparse}
	\end{minipage}%
	\begin{minipage}[c]{.5\textwidth}
		\centering
		\includegraphics[width=0.7\linewidth]{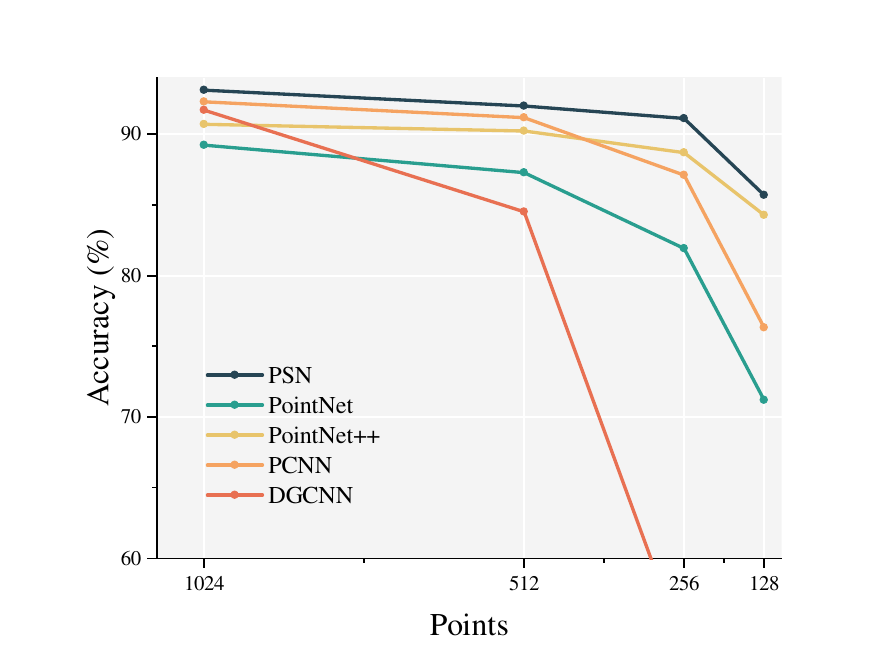}
		\caption{Only evaluate on sparse input.}
		\label{fig:eval_sparse}
	\end{minipage}
\end{figure}
In this section, we analyze the effectiveness of each component including HER, LMIR ClusterFPS and Deep INFOMAX on the ModelNet40 dataset. The results of the ablation study are summarized in Table~\ref{tab:ablation}.

We set two baselines: A and D. Model A is the original implementation of PointNet++~\cite{qi2017pointnetplusplus}, and Model D only has new architecture and ClusterFPS without NEFL. Compared to B, Model D can obtain a $0.5\%$ improvement on mean instance accuracy with fewer input points. When we combine HER with PointNet++ (model C), there is a great improvement in both mean class and instance accuracy. Model E shows the same improvement results on accuracy ($91.1\%$ and $92.8\%$). 

Furthermore, we deploy LMIR into model E to get model H. LMIR shows incremental improvement on accuracy compared to model E. To be fair, we replace the LMIR module in model H with the local mutual information loss (Equation~\ref{Eq:js}) from Deep INFOMAX~\cite{hjelm2019learning} to get model F. Compared to model E without a local mutual information loss, model F fails to promote accuracy and decreases by $0.2\%$ on mean class accuracy. In addition, there is a great gap between models F and H. Thus, we conclude that our LMIR has better performance than Deep INFOMAX on supervised learning. We also replace ClusterFPS with FPS to get model G which has similar performance to model H. Thus, We can draw the conclusion that ClusterFPS may not have the same good sampling uniformity as FPS, but because of the additional cluster information, ClusterFPS still has better performance and faster speed.

\subsection{Robustness for Sparser Input}

To further verify the robustness of the PSN model, we set two robustness experiments: training and evaluating on sparse input (i.e., 1024, 512, 256, 128, 64 and 32), and training on 1024 points but evaluating on sparse input (i.e., 1024, 512, 256 and 128). We compare our method using PointNet~\cite{qi2016pointnet}, PointNet++~\cite{qi2017pointnetplusplus}, SpiderCNN~\cite{xu2018spidercnn}, PCNN~\cite{Matan2018pcnn} and DGCNN~\cite{dgcnn}.

As shown in Figure~\ref{fig:train_sparse}, we can see PSN is very robust to take sparse points for training and evaluating, and PSN drops less than $1\%$, from 1024 points to 256. In each density, PSN outperforms the other methods and still achieves an $89.3\%$ accuracy with 32 points input.

Following~\cite{qi2016pointnet,qi2017pointnetplusplus,Matan2018pcnn,dgcnn}, we train our PSN on an invariant point and evaluate on different densities. As Figure~\ref{fig:eval_sparse} shows, PSN has enough robustness to process sparse points randomly sampled from 1024 points. The experimental results demonstrated that HER and LMIR can help models improve the robustness for a variety of input numbers.

\subsection{Speed Study}

We test the robustness of ClusterFPS by gradually increasing the number of input points from 1024 to 1,000,000. First, we visualize the results of FPS and ClusterFPS. 
We run FPS and ClusterFPS to downsample a point cloud with a batch size of 1. As shown in Table~\ref{tab:time}, ClusterFPS outperforms FPS for every input size. Remarkably, ClusterFPS can downsample one million points to its $10 \%$ size in only 1.53s, achieving up to a $15\times$ speedup over FPS. This shows the dominating capability of our model in processing large-scale point clouds.

\section{Conclusion}
\begin{table}
	\centering
	\tabcolsep=0.12cm
	\begin{tabular}{ccccc}
		\hline
		Num. of Input & 1024 & 4096 & 10000 & 1,000,000 \\
		Num. of Output & 512 & 1024 & 4096 & 100,000 \\
		\hline 
		FPS & 95.1 ms & 187 ms & 798 ms & 230 s \\
		ClusterFPS & 16.3 ms & 32.7 ms & 108 ms & 1.53 s \\
		\hline
	\end{tabular}
	\caption{Downsample speed demonstration of FPS and ClusterFPS. We implement two algorithms in Pytorch with batch size of 1.}
	\label{tab:time}
\end{table}
In this paper, we proposed PointShuffleNet (PSN) for efficient point cloud analysis. PSN showed great promise in point cloud classification and segmentation by introducing the non-Euclidean feature learner (NEFL), which contains a homotopy equivalence relation (HER) based on topology math and a local mutual information regularizer (LMIR). HER introduces the homotopy equivalence transform to our model, makes the neural network to learn the data distribution by homotopy equivalence and extends the generalization. To cut off the trivial homotopy equivalence relation that leads to bad generation, LMIR regularizes the neural network to focus on the non-trivial paths rather than trivial paths by maximizing the mutual information between the original features and HER transformed features. To further improve the efficiency, we proposed a point sampling algorithm named ClusterFPS that is based on the farthest point sampling algorithm. ClusterFPS splits point clouds into several clusters and deploys FPS on each cluster to run parallel computations on the GPUs. PSN achieves state-of-the-art accuracy on classification and achieves great balance between speed and performance on various tasks.

\bibliographystyle{unsrt}  
\bibliography{egbib}  

\newpage
\title{Supplementary Material for PointShuffleNet: Learning Non-Euclidean Features with Homotopy Equivalence and Mutual Information}

\maketitle
\appendix

\begin{abstract}
   This supplementary material is organized as follows:
\begin{itemize}
	\item Section~\ref{sec:glossary} illustrates our math concepts used in our paper through the glossary.
	\item Figure~\ref{fig:psncm} and Figure~\ref{fig:pn2cm} presents the confusion matrices of PointShuffleNet and PointNet++.
	\item Figure~\ref{fig:visio-s3disvis} shows indoor scenes visualization examples segmented by PointShuffleNet on S3DIS dataset.
\end{itemize}
\end{abstract}

\section{Glossary}
\label{sec:glossary}
\begin{itemize}
	\item \textbf{Homotopy Equivalence Relation.} The key concept of homotopy is a continuously deformation between two continuous functions in a topological space. In other words, these functions are said to be equivalent in topological math. 
	We can use homotopy equivalence relation to help neural networks to have better generalization.
	In fact, homotopy equivalence relation can be considered as a principle of feature augmentation in which the augmented features may share a continuous closure subspaces with the original features. We use module shuffle as our deformation method rather than single layer perception for its randomness and high efficiency.
	\item \textbf{Local Mutual Information Regularizer.} We can not control what kind of data Homotopy Equivalence Relation can transform the original data into. 
	However, there is not a mathematical theory to drop unreasonable deformation path in topological.
	 So we propose a regularizer based on mutual information to cut off trivial paths (the unreasonable deformation paths) implicitly named Local Mutual Information Regularizer. We use $f$-GAN~\cite{nowozin2016fgan} to estimate the mutual information between the transformed feature and the original one. So we insert multiple discriminators into networks and utilize different segments of the backbone as generators to form multiple $f$-GAN inside a network. So we can distinguish whether the feature is from a non-trivial path (the reasonable deformation) through mutual information and guides the model to learn the more similar shuffled features and punished the less similar shuffled features which are likely to share less mutual information with the original features.
	\item \textbf{Continuous Closure Subspaces.} The continuous closure subspace is a space that accurately describes the distribution of data, and a class has and only one subspace. But there are infinite decision boundaries between sets. However, we can use the continuous closure subspace to partition the corresponding class and other classes as a decision boundary.
	\item \textbf{Module Shuffle.} It is a shuffle in which a deck of $kn$ cards is divided into $k$ halves which contains $n$ cards. Module shuffle is also called as riffle shuffle or the Faro shuffle if $k=2$. It has been proved that module shuffle is a homotopy equivalence transformation by \cite{Topology2000Anderson}.
	\item \textbf{Trivial Path.} A deformation path connects several point clouds which belong to different class and has intersection with multiple closure subspaces. The trivial path may lead the neural networks to bad generalization. 
	\item \textbf{Non-trivial Path.} A deformation path connects several point clouds which belong to a same class and has no intersection with other closure subspaces. The non-trivial path can be considered as a continuous data augmentation method which interpolates multiple data points inside the closure subspace and clarifies the decision boundaries.
\end{itemize}

\begin{figure}[h]
	\centering
	\includegraphics[width=0.8\linewidth]{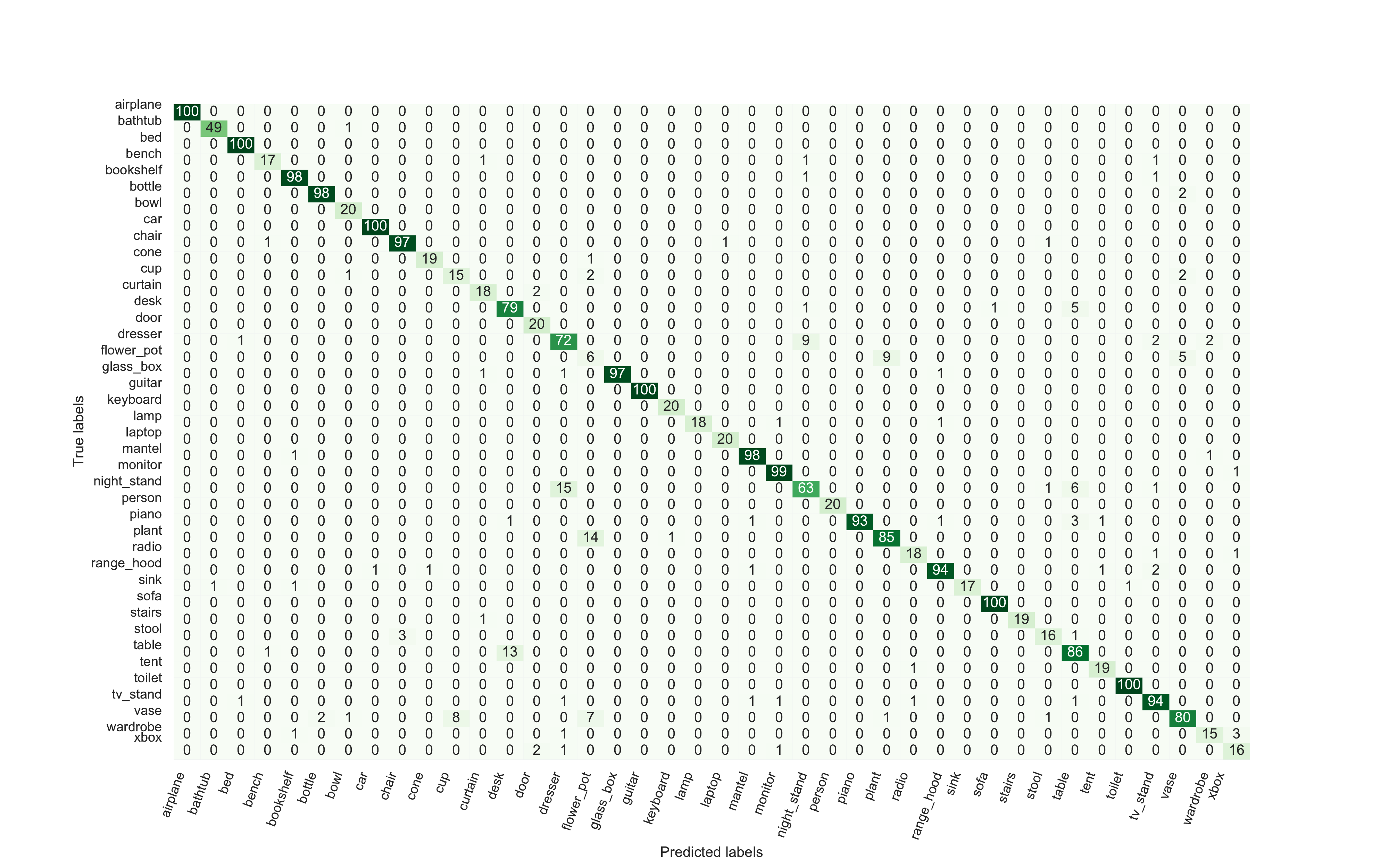}
	\caption{Confusion matrix for PointShuffleNet}
	\label{fig:psncm}
\end{figure}
\begin{figure}[h]
	\centering
	\includegraphics[width=0.8\linewidth]{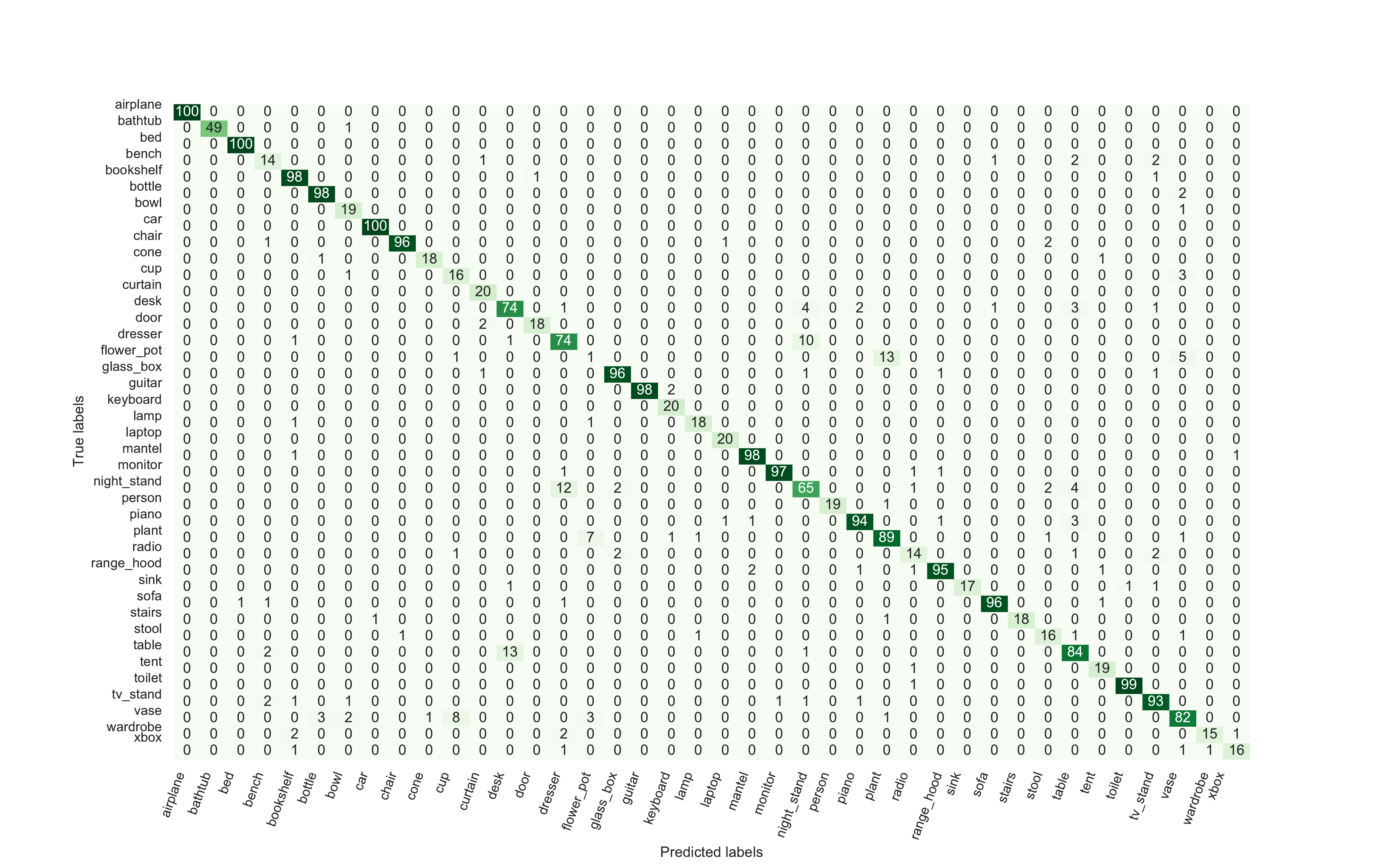}
	\caption{Confusion matrix for PointNet++}
	\label{fig:pn2cm}
\end{figure}

\begin{figure}
	\centering
	\includegraphics[width=0.75\linewidth]{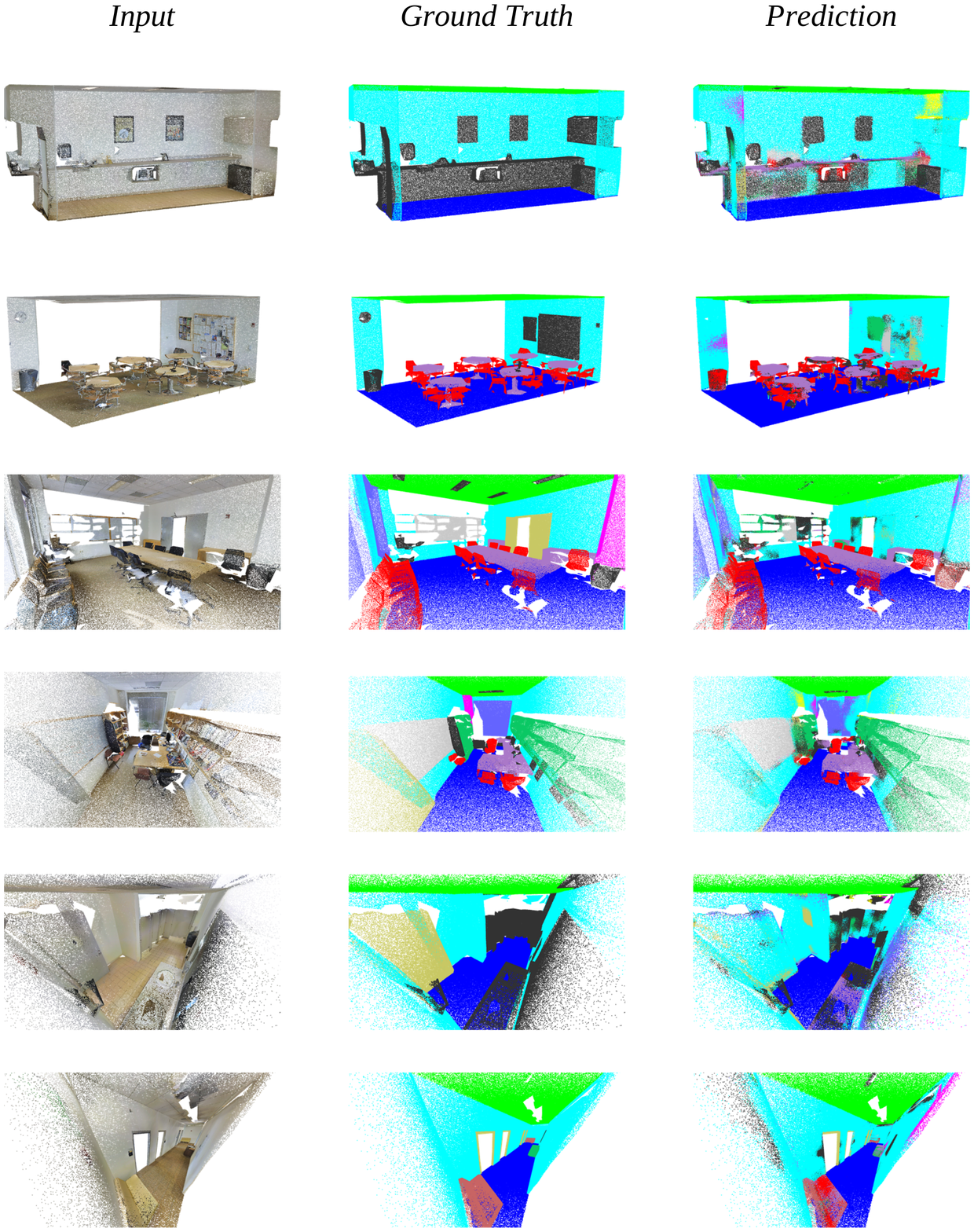}
	\caption{}
	\label{fig:visio-s3disvis}
\end{figure}

\end{document}